\documentclass[letterpaper, 10 pt, conference]{ieeeconf}  

\IEEEoverridecommandlockouts 

\usepackage{graphicx} 
\usepackage{times} 		
\usepackage{amsmath} 	
\usepackage{amssymb}  
\usepackage{algorithm} 
\usepackage{algpseudocode}
\usepackage{pifont}

\usepackage{epsfig}
\usepackage{epstopdf}
\usepackage{multirow}
\usepackage{subfigure}
\usepackage{caption}
\usepackage{lipsum}
\usepackage{gensymb}
\usepackage{cite}

   \def\R{{\mathbb R}}      
     \def\1{{\mathbb I}}

\def\Bx{\mathbf{x}} \def\BX{\mathbf{X}}

\def\Bt{\mathbf{t}} 
\def\Bk{\mathbf{k}} \def\Bp{\mathbf{p}}
\def\BK{\mathbf{K}}

\def\By{\mathbf{y}}

\newtheorem{example}{Example}

\newcommand{\K}{\mathcal{K}}

\renewcommand{\baselinestretch}{0.99} 
\title{\bf
A Nonparametric Motion Flow Model for Human Robot Cooperation
}

\author{Sungjoon Choi, Kyungjae Lee, H. Andy Park, and Songhwai Oh
\thanks{S. Choi, K. Lee and S. Oh are with 
        the Department of Electrical and Computer Engineering and ASRI,  
        Seoul National University, Seoul 151-744, Korea
	(e-mail: \{sungjoon.choi, kyungjae.lee and songhwai.oh\}@cpslab.snu.ac.kr).
	H. Park is with Rethink Robotics, Boston, MA, USA
	(e-mail: apark@rethinkrobotics.com).
}
}
\begin{document}

\maketitle
\thispagestyle{empty}
\pagestyle{empty}

\begin{abstract} 
In this paper, we present a novel nonparametric motion flow model
that effectively describes a motion trajectory of a human 
and its application to human robot cooperation. 
To this end, motion flow similarity measure 
which considers both spatial and temporal properties
of a trajectory is proposed 
by utilizing the mean and variance functions of a Gaussian process. 
We also present a human robot cooperation method 
using the proposed motion flow model.
Given a set of interacting trajectories of two workers,
the underlying reward function of cooperating behaviors
is optimized by using the learned motion description
as an input to the reward function
where a stochastic trajectory optimization method
is used to control a robot. 
The presented human robot cooperation method
is compared with the state-of-the-art algorithm,
which utilizes a mixture of interaction primitives (MIP),
in terms of the RMS error between generated and target trajectories.
While the proposed method shows comparable performance with the MIP
when the full observation of human demonstrations is given, 
it shows superior performance 
with respect to given partial trajectory information. 
\end{abstract}

\section{Introduction}

Traditionally, robots has been deployed to perform 
relatively simple and repetitive tasks in structured environments, 
where robots rarely interacted with humans.
Recently, however, the growing need to reduce human workloads 
and risks and the advances in sensing, actuation, and computing
capabilities have led to the efforts towards human-robot cooperation
\cite{Bauer08}. 
In order for a robot to be effectively deployed at a working space,
where a human and a robot coexist in a close proximity, 
two main issues should be properly handled. 
The first issue is about inferring and recognizing the motion of a 
human co-worker
and the second one is about teaching a robot how to 
act appropriately based on the inferred human motions. 

In this paper, we mainly focus on the first problem 
of recognizing human motions
by finding a mapping from a motion trajectory to a latent vector
that effectively describes the trajectory. 
To this end, a nonparametric motion flow similarity measure
is proposed by measuring the closeness of two trajectories
in terms of both spatial distances and temporal flow directions.
The proposed similarity measure alleviates existing
time-alignment restrictions for computing 
both spatial and temporal distances
using the mean and variance functions of a Gaussian process
\cite{Rasmussen_06}.
This alignment-free property plays an important role in human robot 
cooperation tasks
when it comes to recognizing the human motions given partial 
trajectories, i.e., early recognization. 
Furthermore, we present learning-based human robot cooperation 
based on the proposed motion flow model
combined with inverse reinforcement learning \cite{Choi16_DMRL} 
and stochastic trajectory optimization \cite{Choi16_GRP}.

Both issues of recognizing the human motion and 
planning collaborative robot behavior
have been widely investigated over the years. 
With respect to the first issue of the human motion inference, 
a number of studies have been made
\cite{Li14, Chaandar15, Nikolaidis15}. 
In \cite{Chaandar15},
an interacting multiple model filtering approach is applied to infer
the intended goal position of a human arm. 
\cite{Nikolaidis15} proposed a framework for modeling 
human motions by first clustering human demonstrations
and finding the underlying reward function of each cluster
using inverse reinforcement learning.
While the approach of clustering human demonstrations
is similar to ours, it mainly focused on identifying human
user types.

Once human motion is properly recognized, 
a timely action of a robot should be generated.
In \cite{Mainprice13}, the inferred human motion is used
to predict future movements where robot's motion
is planned by minimizing the penetration cost with
a human. 
A pioneering work on generating interacting behaviors
using interaction primitives (IPs), which is a special type
of dynamic motor primitives~\cite{Paraschos13}, 
was proposed in \cite{Amor14}.
IPs first parameterize trajectories of both the human hand and
the robot state and the joint distribution 
of both parameters is modeled by a Gaussian distribution. 
Then, the interaction between a human and a robot is modeled by a
conditional Gaussian model in the trajectory parameter space.
The interaction primitives are extended to 
a mixture of interaction primitives (MIP) \cite{Ewerton15}
using a Gaussian mixture model and successfully modeled multiple
collaboration tasks between a human and robot.

The main contribution of this paper is to present a 
nonparametric motion flow model that can 
effectively map a trajectory to a latent vector 
considering both spatial and temporal aspects of a trajectory. 
Time-alignment as well as temporal adjustment, e.g., 
dynamic time warping, are not required for the proposed model.
Furthermore, we present a human robot cooperation algorithm
based on the proposed motion flow model
and demonstrated its performance in both simulated and
real-world environments.


\section{Related Work} \label{sec:rel}

Recently, a number of human motion descriptors have been 
proposed in both fields of robotics and computer vision. 
They can be grouped into four different categories based on 
two different criteria.
The first criterion is the source of information: 
a sequence of images or a sequence of positions.
The second criterion is the ability to cope with partially observed trajectories:
time-aligned and alignment-free. 
While some tasks will require explicit time-alignment of trajectories, 
e.g., tasks with timing restrictions or sequential operations,
with respect to human robot collaborations, 
alignment-free descriptors have advantages in that it enables
more natural and rapid interactions between a human and robot.

In \cite{WuL08,yang2011mixed,ShaoL15, HardingE04}, 
a sequence of positions of a human hand or an end effector 
of a robot is given as motion descriptors,
which is often referred to as motion trajectories. 
Geometric invariant descriptions such as curvature and torsion
of a motion trajectory are proposed in 
\cite{WuL08,yang2011mixed,ShaoL15}.
In \cite{WuL08}, curvature, torsion and their first order derivatives 
with respect to arc-length parameters represent a motion description.
In \cite{ShaoL15}, integral invariant features are proposed. 
\cite{HardingE04} proposed self similarity motion descriptions 
by constructing a self similarity matrix based on 
sigmoid distances between all pairs of points in a motion trajectory.
\cite{WuL08,yang2011mixed,ShaoL15, HardingE04} require 
additional dynamic time warping \cite{PaliwalAS82}
to handle the time alignment problem. 
On the contrary, in \cite{HardingE04}, the feature for a motion
trajectory are coefficients of the Fourier transform of the
trajectory, alleviating the time alignment problem.
\cite{muench2014generalized} utilized
the relative joint angles of each skeleton of a motion trajectory
using the generalized Hough transform. 
The time invariant property is achieved as each instance of a motion
trajectory votes for the motion class of the full trajectory.

In computer vision, spatial and temporal motion descriptions 
based on image information have been widely investigated \cite{EfrosBMM03,LinJD09,FathiM08,NieblesWF08,OikonomopoulosPP09}.
\cite{EfrosBMM03} proposed four channel motion descriptors
based on optical flow information and an associated similarity measure 
computed from the normalized correlation between two optical flows.
In \cite{LinJD09}, a representative set of action prototypes is learned 
by training a binary tree using a motion description proposed in \cite{EfrosBMM03}.
\cite{NieblesWF08} presented a motion descriptor using 
the brightness gradients of space time interest points.
In \cite{OikonomopoulosPP09}, the motion description is extracted 
by fitting B-spline surface on detected spatiotemporal salient points 
in image sequences.

\begin{table}[]
\centering
\caption{Classification of motion description algorithms.}
\label{my-label}
\begin{tabular}{|l|l|l|}
\hline
 & Time-aligned   & Alignment-free \\ \hline
Image-based   
&  \cite{FathiM08, LinJD09, EfrosBMM03, SamantaC14, NieblesWF08} 
&  \\ \hline
Trajectory-based
&  \cite{WuL08, yang2011mixed, ShaoL15, HardingE04}             
&  \cite{HardingE04, muench2014generalized}, ours                             
\\ \hline
\end{tabular}
\end{table}

\section{Preliminaries} \label{sec:prel}

In this section, we briefly review density matching reward learning (DMRL)
\cite{Choi16_DMRL} and Gaussian random paths (GRPs) \cite{Choi16_GRP}. 
The DMRL is used to learn the reward function of a robot 
whose input is a motion descriptor of a human co-worker trained 
from the proposed motion flow model for human robot cooperation. 
Once the reward function is learned properly, GRPs are used
to generate an appropriate trajectory of the arm of a collaborating robot. 

\subsection{Density Matching Reward Learning} \label{subsec:dmrl}

Density matching reward learning (DMRL) is a model-free
reward learning algorithm~\cite{Choi16_DMRL},
which finds the underlying reward function given expert's
demonstrations.
Suppose that the estimated density $\hat{\mu}(s,a)$ of state $s$ and
action $a$ is given, DMRL finds a reward function $R(\cdot)$ as follows:
\begin{equation} 
\begin{aligned} \label{eqn:dmrl}
	\underset{R}{\text{maximize}} 
	&& V(R) & = \langle \hat{\mu}, R \rangle \\
	\textrm{subject to} 
	&& \| R \|_2 & \le 1, \\
\end{aligned}
\end{equation}
where the norm ball constraint $\| R \|_2 \le 1$ 
is introduced to handle the scale ambiguity in the 
reward function \cite{ng2000algorithms} and $\langle \hat{\mu}, R \rangle 
= \int_{\mathcal{S} \times \mathcal{A}} \hat{\mu}(s,a)R(s,a) \, ds \, da$. 
For notational simplicity, we will denote $x$ as a state-action pair,
i.e., $x = (s, \, a).$

To model a nonlinear reward function, 
kernel DMRL is proposed by assuming that the reward function
has the following form:
\begin{equation}
	\tilde{R}(x) = 
	\sum_{i=1}^{N_U} \alpha_i k(x, \, {x^U_{i}}),
\end{equation}
where $\{ x^U_i \}_{i=1}^{N_U}$ is a set 
of $N_U$ inducing points,
$k(\cdot, \cdot)$ is a kernel function, and
$\alpha \in \R^{N_U}$ determines the shape 
of the reward function. 
Then, we can reformulate (\ref{eqn:dmrl}) as the following 
unconstrained optimization:
\begin{equation}
\begin{aligned} \label{eqn:dmrl2}
	\underset{\tilde{R}}{\text{maximize}} 
	&&  \tilde{V} = \sum_{\forall x \, \in U} \hat{\mu}(x)\tilde{R}(x) 
		- \frac{\lambda}{2} \| \tilde{R} \|_{\mathcal{H}}^2, \\
\end{aligned}
\end{equation}
where 
$\lambda$ controls the smoothness of the reward function, 
$U = \{ x^U_i \}_{i=1}^{N_U}$ is a set 
of inducing points, and
$\| \tilde{R} \|_{\mathcal{H}}^2$ is the squared Hilbert norm,
which is often used as a regularizer for kernel machines.

\subsection{Gaussian Random Paths} \label{subsec:grp}

A Gaussian random path \cite{Choi16_GRP} defines a distribution over
smooth paths, where a path $\Bp$ is defined as a vector-valued function 
and the domain is a time interval $\mathcal{I} \subseteq \R $ and the
range is a set of locations.  
Given a kernel function $k(t, \, t')$ 
and a set of $M$ anchoring locations 
$D_{a} = (\Bt_{a}, \Bx_{a}) = \{ (t_i, \, x_i) \}_{i=1}^M$, 
a Gaussian random path $\mathcal{P}$ 
with a sequence of $T$ test time indices
$\Bt_{test} = \{ t_i \}_{i=1}^T$
is specified by a mean path $\mu_{\mathcal{P}}$
and a covariance matrix $K_{\mathcal{P}}$, i.e.,
	$\mathcal{P} \sim \mathcal{N}(\mu_{\mathcal{P}}, \, K_{\mathcal{P}})$,
where 
\begin{align} 
	\mu_{\mathcal{P}} 
	&=
	\Bk( \Bt_{a}, \, \Bt_{test})^T (\BK_{a} + \sigma^2_wI)^{-1}\Bx_{a},  
	\label{eqn:grp_mean}
	\\
	\K_{\mathcal{P}}
	&=
	\BK_{test} - \Bk(\Bt_{a}, \, \Bt_{test})^T 
	(\BK_{a} + \sigma^2_wI)^{-1} \Bk(\Bt_{a}, \, \Bt_{test}), \nonumber
\end{align}	
$\Bk(\Bt_{test}, \Bt_{a}) \in \R^{T \times M}$ 
is a kernel matrix of test time indices and 
anchoring time indices, 
$\BK_{a} = \BK(\Bt_{a}, \Bt_{a}) \in \R^{M \times M}$ 
is a kernel matrix of anchoring time indices, 
and $\BK_{test} = \BK(\Bt_{test}, \Bt_{test}) \in \R^{T \times T}$ 
is a kernel matrix of test time indices. 
In this paper, a squared exponential kernel function is used as the kernel function.
An $\epsilon$ run-up method was further proposed to add a directional
constraint on the path distribution \cite{Choi16_GRP}. 

Gaussian random paths are used for two purposes.
First, as the GRP defines a distribution over paths, diverse smooth
paths interpolating a set of anchoring points can be efficiently
sampled.  
While existing trajectory optimization methods, such as STOMP
\cite{Kalakrishnan_11} and CHOMP \cite{Ratliff_09}, can be also used, 
we have found that trajectory optimization with GRPs is faster and
produces a higher quility solution on problems that are sensitive to the
choice of an initial trajectory \cite{Choi16_GRP}. 
Secondly, the mean path of GRPs can be used in a preprocessing step
of the proposed method for smoothing and computing time derivatives. 

\section{Nonparametric Motion Flow Model} \label{sec:flow}

In this section, we propose a nonparametric motion flow model
for inferring and recognizing human motion trajectories,
where a motion flow is defined as a mapping between 
positions of a trajectory to its derivatives.
Given motion trajectories of a human, we first cluster
trajectories using spectral clustering with the proposed motion flow 
similarity measure, where trajectories of each cluster 
are used to train a nonparametric motion flow model. 
Finally, the motion description of a trajectory
is achieved by computing
how close the trajectory is to each motion flow model in terms of
the motion flow similarity measure. 

In the following sections, we will focus on the position of the right hand
of a human, however, higher dimensional inputs, e.g., 
joint positions of a right arm, can be also used. 
Suppose that $\Bx \in \R^3$ and $\dot{\Bx} \in \R^3$ are the position
and velocity of a right hand. 
Then, the flow function $f(\cdot)$ maps $\Bx$ to $\dot{\Bx}$, i.e., 
$f: \Bx \mapsto \dot{\Bx}$.

For notational simplicity, we will denote $\xi$
as a trajectory containing both positions and velocities, i.e., 
$\xi = \{ (\Bx_k, \dot{\Bx}_k ) \}_{k=1}^{L}$,
where $L$ is the length of a trajectory
and 
$\Bx_{i, t}$ and $\dot{\Bx}_{i, t}$
are the position and velocity at time $t$ for the $i$th trajectory $\xi_i$, 
respectively.

\subsection{Motion Flow Similarity Measure}

Suppose that two trajectories $\xi_i$ and $\xi_j$ are given. 
We present a nonparametric motion flow similarity measure
based on both spatial and temporal aspects of trajectories,
where the spatial similarity
indicates how spatially close $\xi_i$ and $\xi_j$ are and 
the temporal similarity represents how much velocities of 
$\xi_i$ and $\xi_j$ are aligned. 

Given two trajectories $\xi_i$ and $\xi_j$, 
a motion flow similarity measure $d(\xi_i, \, \xi_j)$
of $\xi_i$ and $\xi_j$ is defined as:
\begin{equation} \label{eqn:metric}
	{d}(\xi_i ; \, \xi_j) = 
	\frac{1}{L}\sum_{t=1}^L
			\left(
			d_{cos}(
				\dot{\Bx}_{i,t}, \, \hat{\mu}_j(\Bx_{i,t})
				)
			+ \hat{\sigma}_j^2(\Bx_{i,t})
			\right),
\end{equation}
where
$\hat{\mu}_j(\cdot)$ and $\hat{\sigma}_j^2(\cdot)$ are the 
Gaussian process mean function and variance function
trained with input-output pairs $(\Bx_{j,t}, \, \dot{\Bx}_{j,t})$ from
$\xi_j = \{ (\Bx_{j,t}, \, \dot{\Bx}_{j,t}) \}_{t=1}^{L}$.
$\hat{\mu}_j(\cdot)$ and $\hat{\sigma}^2_j(\cdot)$ are
defined as follows:
\begin{equation}
	\hat{\mu}_j(\Bx_{i,t}) =  k(\Bx_{i,t})^T (k(\mathbf{X}, \mathbf{X})
	+ \sigma^2_wI)^{-1} \mathbf{y}
\end{equation}
and
\begin{equation}
	\hat{\sigma}^2_j(\Bx_{i,t}) = k_{\star} - k(\Bx_{i,t})^T	(k(\mathbf{X}, \mathbf{X}) + \sigma^2_wI)^{-1} k(\Bx_{i,t})
\end{equation}
where 
$\BX = \{ \Bx_{j,t} \}_{t=1}^{L} $, 
$\By = \{ \dot{\Bx}_{j,t} \}_{t=1}^{L} $, 
$k_{\star} = k(\Bx_{i,t}, \Bx_{i,t})$,
$k(\Bx_{i,t}) \in \R^n$ is a covariance vector between
the test point $\Bx_{i,t}$ and $L$ input points $\mathbf{X}$,
and $k(\mathbf{X}, \mathbf{X}) \in \R^{L \times L}$ is a
covariance matrix
between $L$ data points $\mathbf{X}$.
Here, $d_{cos}(\cdot, \, \cdot)$ is a cosine distance between two
velocities: 
\begin{equation} \label{eqn:dist}
	d_{cos}(\dot{\Bx}_a, \dot{\Bx}_b) =
	1 - \frac{\dot{\Bx}_a^T\dot{\Bx}_b}
		{\| \dot{\Bx}_a \|_2 \| \dot{\Bx}_b \|_2}.
\end{equation}

Intuitively speaking, (\ref{eqn:metric}) indicates a distance measure 
between two trajectories considering both spatial and temporal properties. 
To be more specific, 
the first term inside the summation in (\ref{eqn:metric}), i.e., 
$\frac{1}{L}\sum_{t=1}^L d_{\cos}(
\dot{\Bx}_{i,t}, \, \hat{\mu}_j(\Bx_{i,t})
)$,
indicates the 
\textit{temporal similarity} between $\xi_i$ and $\xi_j$ as it
represents how the velocities of $\xi_i$ and $\xi_j$ are aligned. 
Similarily, the second term, i.e.,
$\frac{1}{L}\sum_{t=1}^L 
\hat{\sigma}_j^2(\Bx_{i,t})$,
represents the \textit{spatial similarity} between $\xi_i$ and $\xi_j$
since the Gaussian process variance 
increases as the distance between the
test input and training samples
increases, i.e., the predictive variance is high if there are no nearby training samples.
Both similarities go to zeros if $\xi_i=\xi_j$.
We would like to emphasize that 
time-alignments of $\xi_i$ and $\xi_j$ 
are not required as they are not compared in a point-wise manner. 
We would like to note that the proposed motion flow similarity measure
can be applied to any input and output sequences
by treating $\Bx$ as an input and $\dot{\Bx}$
as an output. 

\begin{figure}[!t] \centering
	\includegraphics[width=.95\columnwidth]{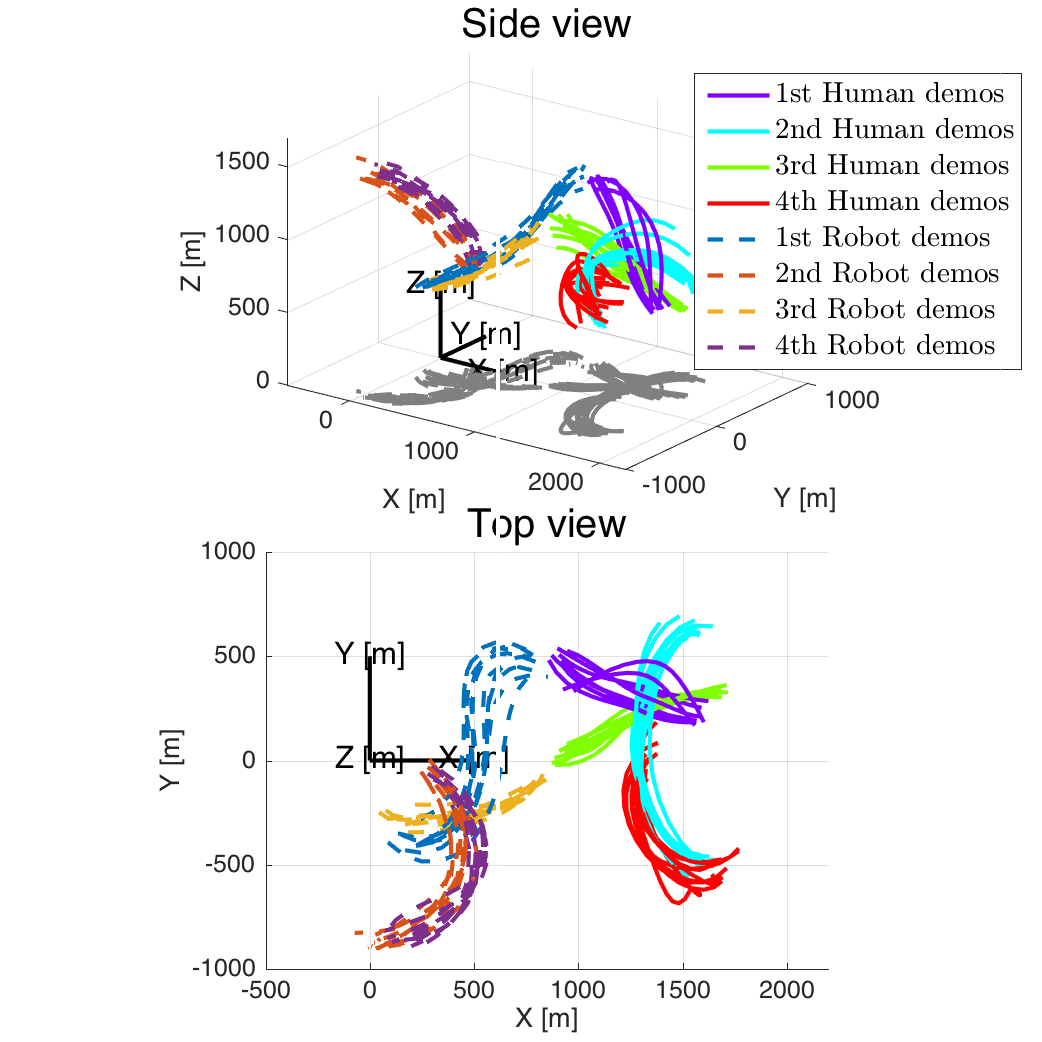} 
	\caption{
		A side view and top-down view of $40$ trajectories
		clustered in different colors.
		}
	\label{fig:clusters}
\end{figure}

\subsection{Training Motion Flow Model} 

Suppose that motion trajectories are given as training data. 
The collected trajectories are clustered using a spectral clustering
algorithm \cite{Ng02spectral}.  
The adjacency matrix $A \in \R^{N \times N}$ is computed 
from the similarity measure (\ref{eqn:metric}), where $N$ is the
number of trajectories. 
In particular, $[\bar{A}]_{i, j} = d(\xi_i ; \xi_j)$
and $A = \frac{1}{2}\ (\bar{A}^T + \bar{A}$)
for ensuring positive definiteness of the adjacency matrix. 
Clustering results of four clusters using 
$40$ training samples are shown in Figure \ref{fig:clusters} 
(see Section~\ref{subsec:data} for information about how samples are collected).

Once trajectories are clustered, a 
motion flow model is trained 
using Gaussian process regression (GPR) \cite{Rasmussen_06}.
In particular, the motion flow function of each cluster
is trained using GPR, where the positions of trajectories
are given as input training data and corresponding 
velocities are given as output training data. 
In other words, given $K$ clusters, we compute
$K$ GP mean functions, $\hat{\mu}_k(\cdot)$, 
and variance functions, $\hat{\sigma}_k^{2}(\cdot)$.
Furthermore, we compute $K$ motion flow similarity 
measures 
\begin{equation} \label{eqn:metric_k}
	{d_k}(\xi_i) = 
	\frac{1}{L}\sum_{t=1}^L
			\left(
			d_{cos}(
				\dot{\Bx}_{i,t}, \, \hat{\mu}_k(\Bx_{i,t})
				)
			+ \hat{\sigma}_k^2(\Bx_{i,t})
			\right).
\end{equation}

\begin{figure}[!t] \centering
	\includegraphics[width=.85\columnwidth]{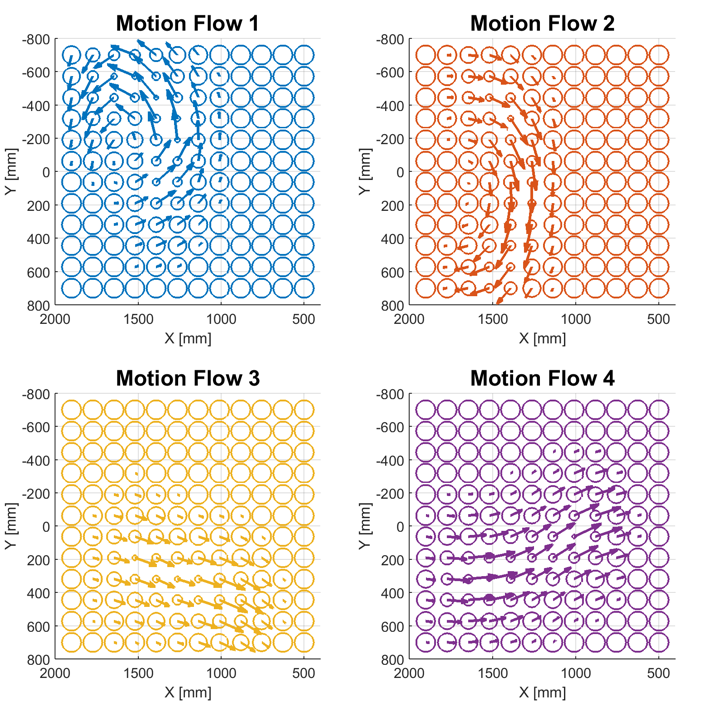} 	
	\caption{
		A top-down view of nonparametric motion flow models. 
		A motion flow is shown with an arrow and the size of a circle
		indicates the predicted variance $\hat{\sigma}^2(\cdot)$ at each grid. 
		}
	\label{fig:gpim}
\end{figure}

The motion flow model trained with each cluster is illustrated in 
Figuere \ref{fig:gpim}, where the arrow indicates the 
Gaussian process mean functions $\hat{\mu}(\cdot)$ 
and the size of the circle is proportional to the predictive variance 
$\hat{\sigma}^2(\cdot)$.
We can easily see that the direction of the arrow 
coincides with the direction of trajectories in each cluster
while the predictive variances of GP increase as we move away from
the mean path. 

\subsection{Motion Flow Description Inference}
\label{subsec:inf}

Finally, the trained motion flow model is used to extract
motion flow descriptions of a motion trajectory. 
The motion flow descriptions are computed as follows. 
Suppose that the number of clusters is $K$. 
Then the motion flow description $\Bp$ of a motion trajectory $\xi$
is a $K$-dimensional vector 
on a simplex, where $k$-th element 
is proportional to 
$\exp(-d_k(\xi)^2)$, i.e., $
\{\mathbf{p} \in [0,1]^K \, 
| \, p_k \propto \exp(-d_k(\xi)^2) \,\sum_{k=1}^K p_k=1 \}$.
We will denote the mapping between a trajectory $\xi$
and motion descriptor $\Bp$ as $\phi(\cdot)$, i.e., 
$\phi: \xi \mapsto \Bp$.

The motion flow description $\Bp$ representing a trajectory
is used to model the reward function of a robot given 
interacting demonstrations consisting of motion trajectories
of a human and robot. 
As the motion flow description uses flow functions trained 
with a GP, time alignment of trajectories is not required. 
This alignment-free property plays a significant role
in early recognition of the human motion in upcoming
experiments shown in Section \ref{sec:exp}.

\section{Experiments} \label{sec:exp}

In this section, the performance of the proposed
nonparametric motion flow model is extensively validated
through human robot cooperation. 
The proposed method is compared with
a mixture of interaction primitives (MIP)~\cite{Ewerton15}
in terms of root mean squared (RMS) errors
between predicted and target trajectories in the test set
under different scenarios. 

\subsection{Collecting Interacting Demonstrations}
\label{subsec:data}

Before presenting the experimental results of the proposed method, 
let us introduce the interacting demonstration dataset used 
throughout the experiments. 
We assume that a human and a robot are collaboratively 
arranging the space in a close proximity under four different modes:
\begin{enumerate}
	\item Center-hand-over: If a human hands over 
	an object in the middle, a robot reach the object by 
	stretching its end-effector to the middle. 
	\item Right-hand-over: If a human hands over 
	an object in the right, a robot reach the object by 
	stretching its end-effector to the right.
	\item Right-swipe: If a human swipes his hand 
	to the right, a robot operates its own arranging 
	movement of raising its end-effector to the top right. 
	\item Left-swipe: If a human swipes his hand to the left,
	a robot operates its own arranging movement of raising 
	its end-effector to the top right.  
\end{enumerate}

\begin{figure*}[t!]
  \centering
  \includegraphics[width=1.99\columnwidth]{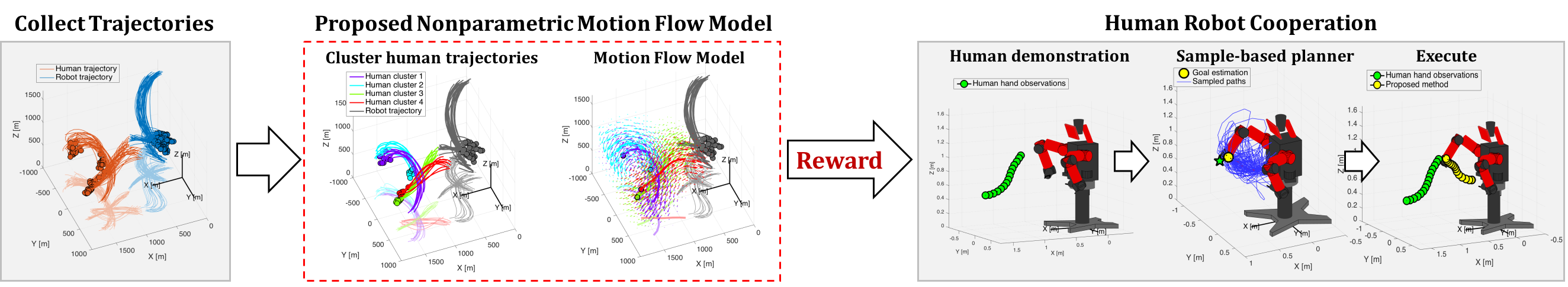}
  \caption{
    An overview of the human robot cooperation algorithm based
    on the proposed motion flow model. 
    Given interacting demonstrations between a human and robot, a 
    motion flow model is trained by first clustering motion
    trajectories of a human and compute motion flow similarity
    measures of each cluster. 
    Then, the underlying reward function of a robot is optimized
    and further used in the execution phase using a sampling-based
    trajectory optimizer. 
  }
  \label{fig:overview}
\end{figure*}

We have collected interacting demonstrations between two persons.
In particular, we have recorded three dimensional positions of
right hands of two persons using a Vicon motion capture system
at $10$Hz.  
One person performs the role of a robot considering the working range
of a robot.\footnote{A Baxter robot from Rethink Robotics 
is considered in this paper.} 
We have collected $20$ demonstrations for each mode and the collected
demonstrations are further divided into two sets: train and test sets. 
The collected trajectories are shown in the left of
Figure~\ref{fig:overview}.

%
\subsection{Implementing Human Robot Cooperation Algorithm} \label{subsec:}

Once a nonparametric motion flow model is  
trained from demonstrations of a human,
the reward function of the robot
that explains the interacting behavior is optimized using
inverse reinforcement learning. 
This indirect way of modeling the interacting behaviors 
of the robot is particularly important as it allows 
incorporating additional considerations, such as collision avoidance,
to the optimized reward function. 
The reward function $r$ is a function of both motion flow descriptions
of a human $\phi(\xi^O)$
and the end-effector position of a robot $\Bx^R$, 
i.e., $r(\phi(\xi^O), \, \Bx^R)$.

Suppose that a set of $N$ interacting demonstrations 
of both a human and a robot, 
$\mathcal{D} = \{ (\xi^H_i, \, \xi^R_i) \}_{i=1}^N$,
is given and the number of clusters in the motion flow model
is $K$. 
For notational simplicity, we assume that the length of 
each trajectory is $L$. 
Let $\xi^H_{i, 1:t}$ be the $i$th trajectory of a human 
from time $1$ to $t$, where $1 \le t \le L$.
We first extract motion flow features $\Phi$
from human demonstrations at each time step, 
i.e., $\Phi = \{ \phi(\xi^H_{i, 1:t}) \, | \, t=1, ..., L\}_{i=1}^N$
along with the position of the end-effector of a robot. 
In other words, we extract $N \times L$ pairs
of $K$ dimensional motion flow features 
and corresponding end-effector positions
from $\mathcal{D}$, which will be used as a train set.
Density matching reward learning \cite{Choi16_DMRL} 
is used to optimize the underlying reward function
of a robot given recognized human motions. 
However, other inverse reinforcement learning algorithms,
e.g., Gaussian process inverse reinforcement learning
\cite{levine2011nonlinear}
or continuous inverse optimal control \cite{Levine_12},
can also be used. 

Furthermore, we similarly define the reward of a trajectory
of the end-effector of a robot by averaging the reward
of each position in the trajectory. 
Let $\xi^R$ be the end-effector trajectory of a robot, then 
the reward of the robot trajectory 
given the observed human trajectory $\xi^O$
is defined as:
\begin{equation} \label{eqn:reward_traj}
	R(\xi^R | \, \xi^O) = \frac{1}{L} \sum_{t=1}^L 
		r(\phi(\xi^O), \, \Bx^R_t).
\end{equation} 

Once the motion of a human is inferred, 
a pertinent trajectory of the right arm of a robot 
is optimized with respect to the reward function.
First, we estimate the final pose and its time derivative
of an end-effector trajectory. 
Suppose that $\mathcal{D} = \{ (\xi^H_i, \, \xi^R_i) \}_{i=1}^N$
is the interacting demonstrations of a human and a robot. 
Then, the final pose of a robot $\mathbf{\hat{x}}^R_L$ 
given an observed human trajectory $\xi^O$
is estimated as: 
\begin{equation} \label{eqn:finalpose}
	\begin{aligned}
		&& \mathbf{\hat{x}}^R_L
		& 
		=
		\sum_{i=1}^{N} \mathbf{x}^R_{i,L}
			\frac{p(\xi^O | \, \xi^H_{i})}
			{
				\sum_{j=1}^N p(\xi^O | \, \xi^H_{j})
			},
	\end{aligned}
\end{equation} 
where $L$ is the length of a trajectory and 
$\mathbf{x}^R_{i,L}$ is the last position of the $i$th 
robot trajectory $\xi_i^R$. 
The final time derivative $\mathbf{\hat{\dot{x}}}^R_L$ 
is similarly estimated by replacing 
$\mathbf{x}^R_{i,L}$ to $\dot{\mathbf{x}}^R_{i,L}$
in (\ref{eqn:finalpose}).

\begin{figure}[t!]
	\centering
	\includegraphics[width=0.8\columnwidth]
		{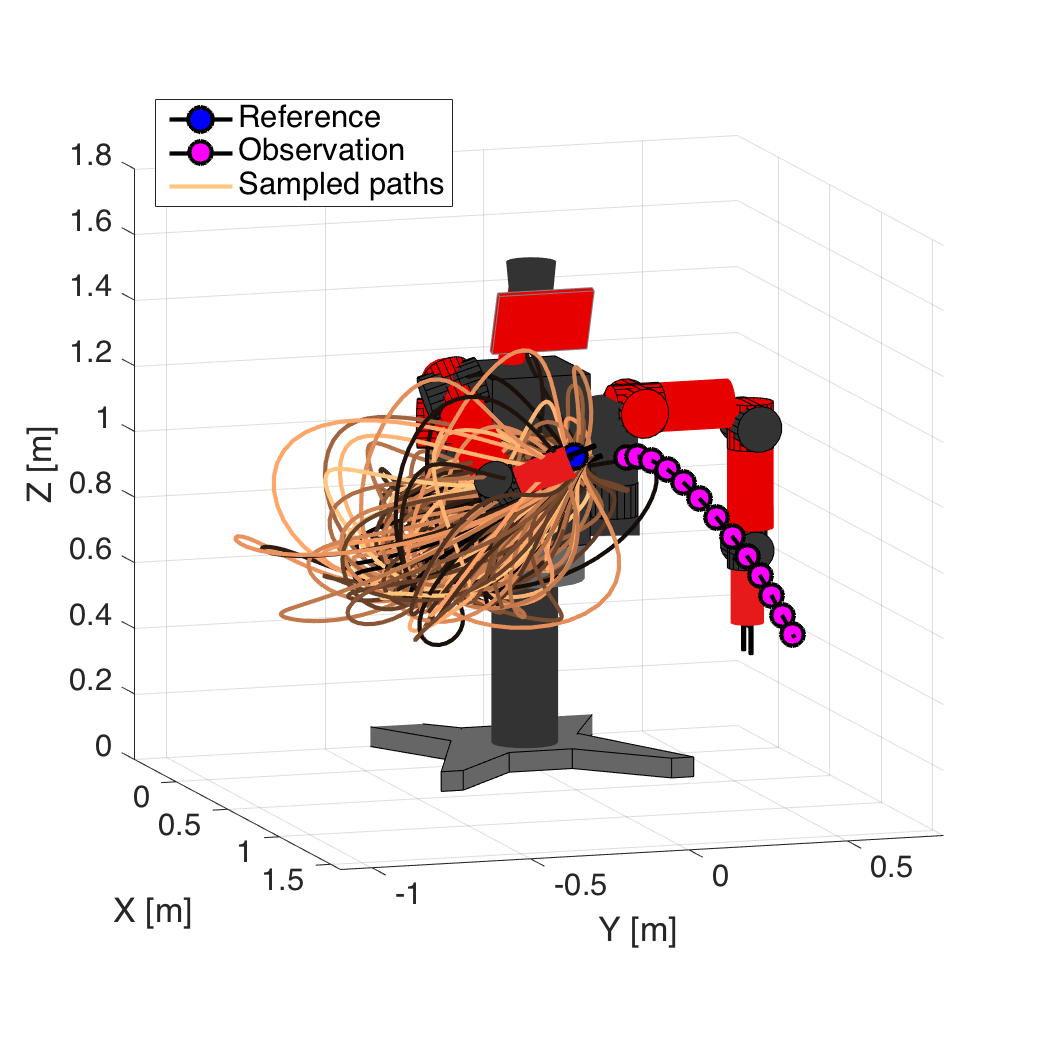}
	\caption{
		Sampled paths for a Baxter robot using
		Gaussian random paths. 
	}
	\label{fig:grp}
\end{figure}

Finally, the interacting trajectory of a robot end-effector,
$\hat{\xi}^R$, is computed as:
\begin{equation} \label{eqn:trajopt}
	\begin{aligned}
		&& \hat{\xi}^R
		& =
		\sum_{i=1}^K \xi_i^R 
			\frac{\exp(R(\xi^R_i|\xi^O))}
			{
				\sum_{j=1}^K \exp(R(\xi^R_j|\xi^O))
			},
	\end{aligned}
\end{equation}
where $R(\xi^R_i|\xi^O)$ is the reward of the trajectory of a robot's
end-effector in (\ref{eqn:reward_traj}) and 
$\xi^R_i$ is the $i$th sampled trajectory 
drawn from a Gaussian random path 
distribution $p_{GRP}(\xi^R)$. 
The anchoring points of the GRP consist of 
$\{(0,\,\Bx^R_t)
, \, (1-\epsilon,\, \hat{\Bx}^R_L - \epsilon \hat{\dot{\Bx}}^R_L)
, \, (1,\,\hat{\Bx}^R_L)\}$, where $\Bx^R_t$
is the current end-effector position and the second anchoring point is
added to incorporate the estimated final heading of a robot, where
$\epsilon$ is set to $0.01$. 
$200$ paths of the end-effector of a Baxter
robot sampled from the GRP are shown
in Figure~\ref{fig:grp}. 
The path sampling is done in the $7$-DoF configuration space
of a Baxter's right arm 
and the three dimensional paths of the end-effector are 
computed using the forward kinematics.
The overall computation took less than 
$100$ms in MATLAB on a $2.2$GHz quad-core processor. 
We will refer the human robot cooperation 
algorithm with the proposed motion flow model
as intention aware apprenticeship learning (IAAL). 
The overview of IAAL is shown in Figure \ref{fig:overview}.  

\subsection{Planning With Partial Observations} \label{subsec:partial}

In this experiment, we validated the early recognition performance of
IAAL compared to MIL
by varying the observation ratio of a human hand trajectory. 
We vary the observation ratio from $0.2$ to $1.0$,
where $0.2$ indicates giving first $20\%$ of 
the human hand trajectory. 
The number of clusters for the IAAL and MIP
are set to five.

\begin{figure}[!t] \centering
	\includegraphics[width=.7\columnwidth]{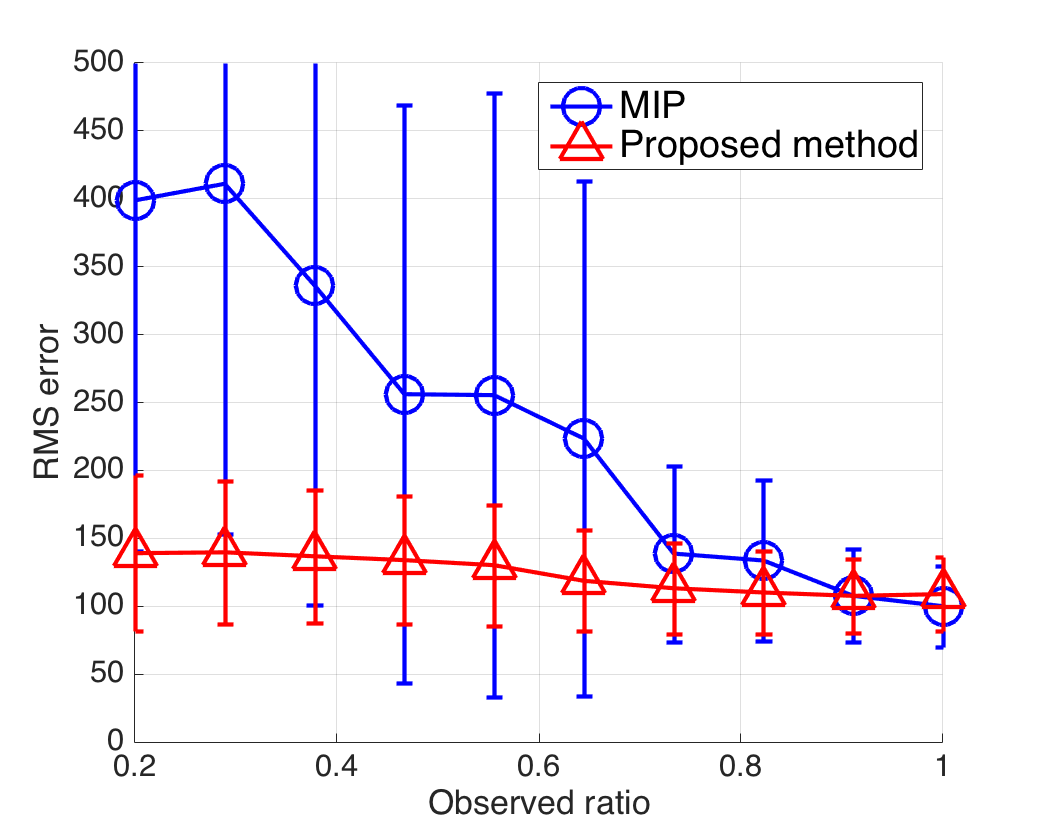} 
	\caption{
		RMS prediction errors of the proposed IAAL and 
		MIP as a function of the observation ratio.
		}
	\label{fig:exp_obsrate}
\end{figure}

The prediction results are shown in Figure~\ref{fig:exp_obsrate}. 
While IAAL shows comparable performance with MIP
given the full observation of a human hand trajectory, 
the prediction performance of MIP degenerates significantly as 
the observation ratio decreases. 
On the contrary, the performance drop of IAAL
is minuscule reflecting the early prediction ability of IAAL. 
This is largely due to the proposed motion flow similarity measure,
which is free from the time alignment of trajectories.

\begin{figure}[!t] \centering
	\subfigure[]{\includegraphics[width=.7\columnwidth]{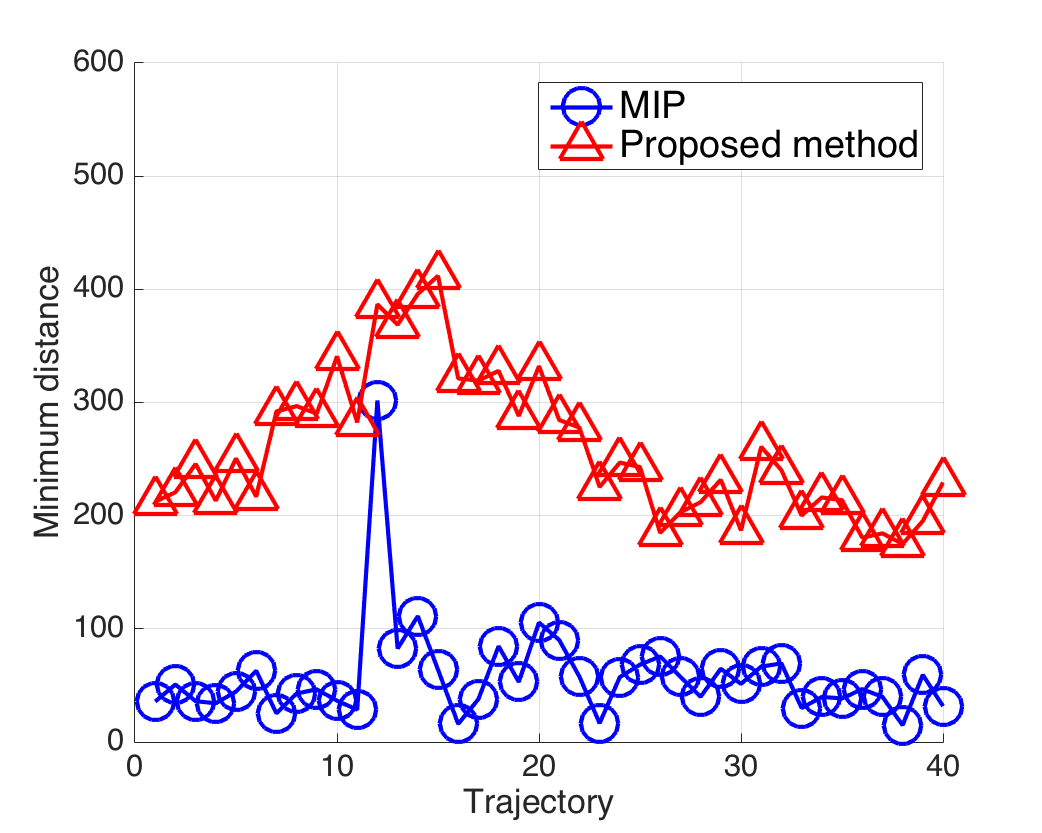} 
	\label{fig:mindists}}
	\subfigure[]{\includegraphics[width=.7\columnwidth]{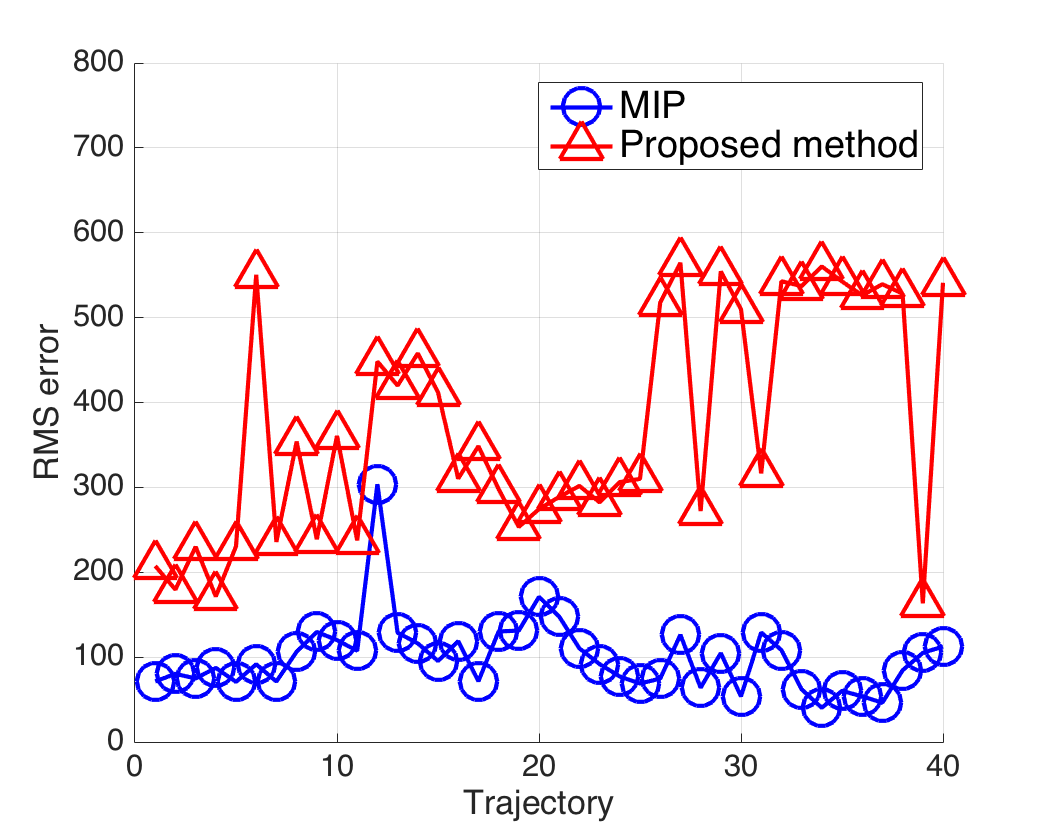} 
	\label{fig:rmsobs}}
	\caption{
		(a) Minimum distances to the obstacle of each run. 
		(b) RMS prediction errors of each run.
		}
\end{figure}

\subsection{Planning With Obstacles}
\label{subsec:obs}

The objective of this experiment is to 
validate the ability of IAAL of incorporating 
additional considerations. 
In particular, an obstacle is assumed to be placed 
in the middle of the target trajectory of the end-effector. 
We consider not only the collision with the end-effector 
of a robot, but also three joints
(wrist, elbow, and shoulder) in the right arm of a Baxter robot. 
This becomes possible as we sample paths in the configuration (joint) 
space of the right arm of a Baxter robot and use forward kinematics 
to compute the trajectories of
four joints (hand, wrist, elbow, and shoulder)
in a three dimensional Cartesian coordinate. 

The following log-barrier function is designed for collision avoidance:  
\begin{equation} \label{eqn:barrier}
	C^{obs}(\xi^R) = -\log \left(\gamma \cdot (d_{min}-\alpha)+\epsilon \right) + \beta,
\end{equation}
where $d_{min}$ is the minimum distance from the 
obstacle to four trajectories of the right arm, i.e., 
right hand, wrist, elbow, and shoulder, 
$\alpha = 100$, $\beta=-12.8$, $\gamma=5.73 \times 10^{-9}$, 
and $\epsilon = 10^{-6}$.

As MIP cannot directly handle additional constraints,
only IAAL is modified by subtracting (\ref{eqn:barrier}) to the reward
function in the trajectory optimization process 
in (\ref{eqn:trajopt}) as follows:
\begin{equation*} \label{eqn:trajobs}
	\begin{aligned}
		&& \hat{\xi}^R
		&& & = 
		\sum_{i=1}^K \xi_i^R 
			\frac{\exp(R(\xi^R_i|\xi^O)- C^{obs}(\xi_i^R))}
			{
				\sum_{j=1}^K \exp(R(\xi^R_j|\xi^O)-C^{obs}(\xi_i^R))
			}.
	\end{aligned}
\end{equation*}

\begin{figure*}[!t] \centering
	\subfigure[]{\includegraphics[width=1.8\columnwidth]{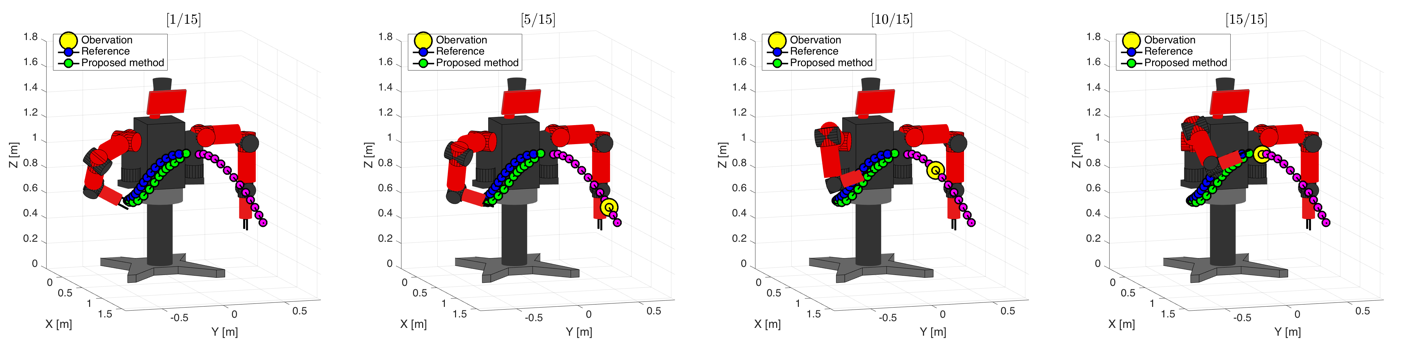} 
	\label{fig:seq_w_obs}}
	\subfigure[]{\includegraphics[width=1.8\columnwidth]{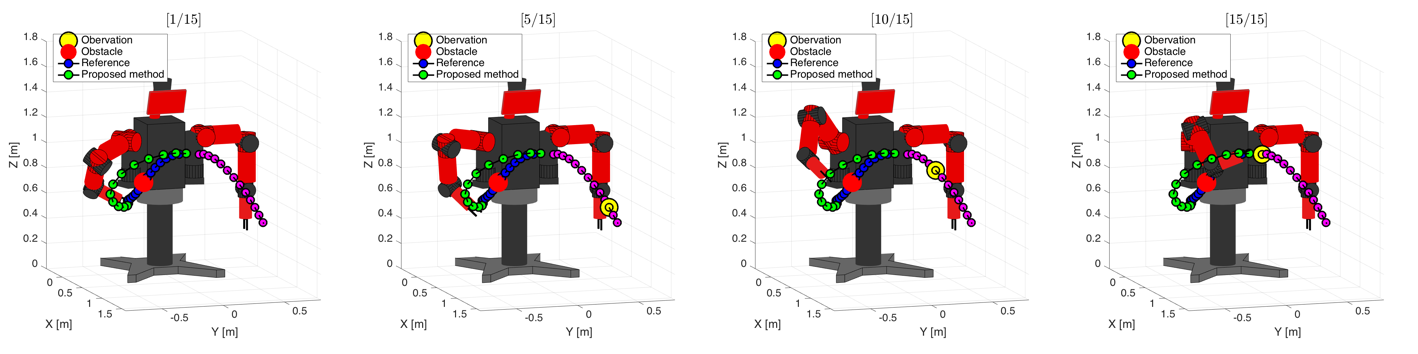} 
	\label{fig:seq_wo_obs}}
	\caption{
		(a) Snapshots of a Baxter robot executing the optimized trajectory using IAAL
		in a free space. 
		(b) Snapshots of a Baxter robot executing the optimized trajectory using IAAL
		with an obstacle shown as a red circle in the middle of the target trajectory. 
		}
	\label{fig:snap_obs}
\end{figure*}

The minimum distance to obstacles 
and the RMS prediction error of $40$ test runs.
are shown in Figure \ref{fig:mindists} and \ref{fig:rmsobs},
respectively. 
While the RMS prediction errors increase as 
the end-effector trajectories of IAAL detour to prevent collision, 
the minimum distance from obstacles to four trajectories
of the right arm (end-effector, wrist, elbow, and shoulder)
exceeds $180$mm for all $40$ test runs. 
Snapshots of a Baxter robot executing hand over tasks with and without
obstacles are shown in Figure \ref{fig:snap_obs}.

\begin{figure}[t!]
	\centering
	\includegraphics[width=0.86\columnwidth]
		{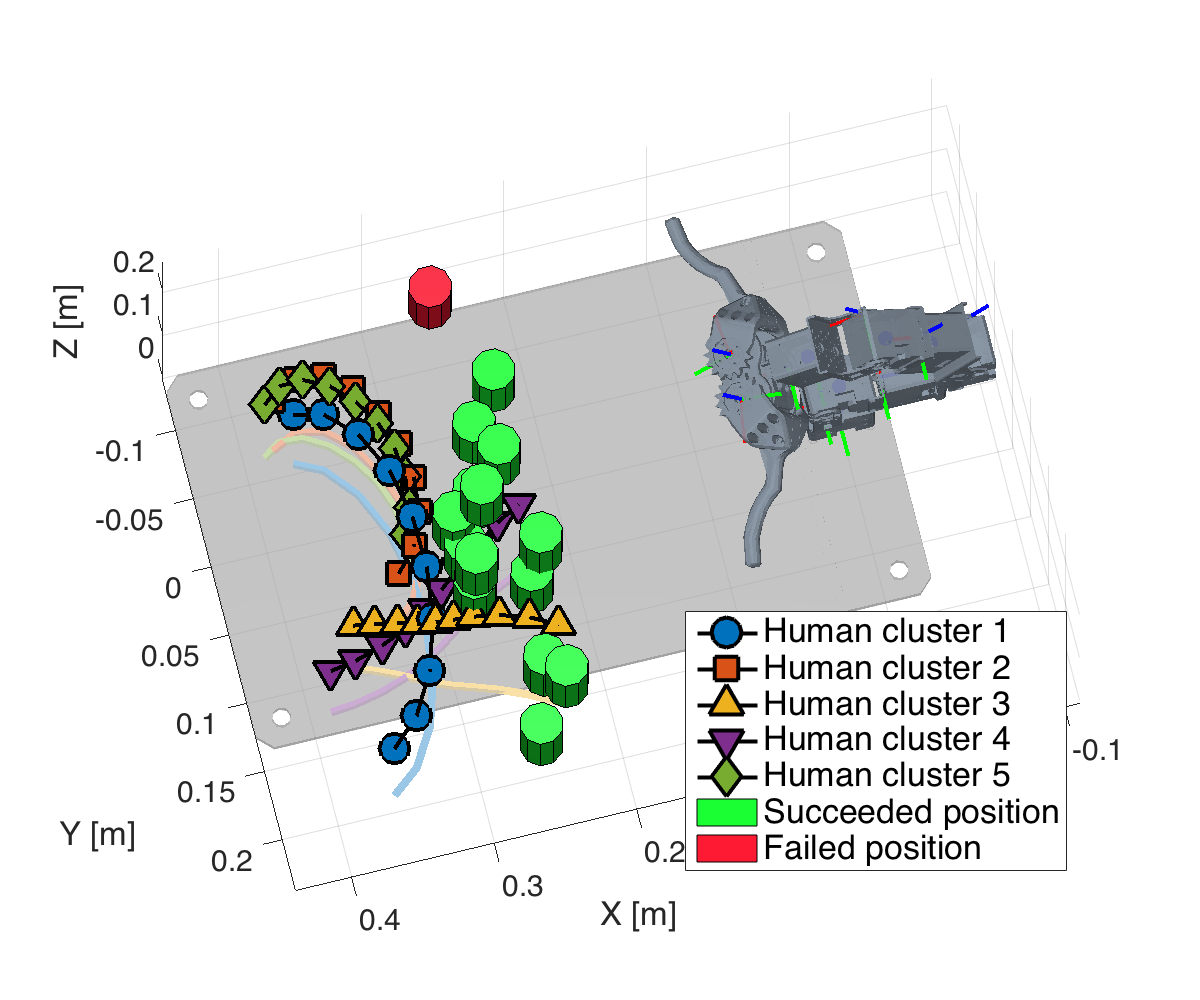}
	\caption{
		Successful and failed final grasped positions of an object are shown with
		green and red cylinders, respectively. 
		The mean trajectory of each cluster is shown with 
		a different marker. 
	}
	\label{fig:grasploc}
\end{figure}

%
\subsection {Manipulator-X Experiments}

In this section, Manipulator-X from Robotis, 
a $7$-DoF manipulator, is used to validate the applicability 
of the proposed algorithm in real-world environments.
The interacting demonstrations in Section \ref{subsec:data}
is also used in this regard, where the positions of demonstrations are 
rescaled to compensate the smaller working range of 
Manipulator-X. 

Once demonstrations are downscaled to fit Manipulator-X, 
the proposed IAAL algorithm is used plan a pertinent action of the
manipulator, where the number of clusters is set to five. 
The human hand position is captured by a Vicon motion capture system
and the manipulator re-plans its trajectory every two seconds 
with respect to the recognized human motion. 
The overall planning procedure took less than $400$ms
in MATLAB on a $2.2$GHz quad-core processor
and the manipulator halts its motion while planning.

We first conducted simple hand-over experiments,
where an additional grasping motion is manually programmed.
The final grasped positions 
along with the average trajectory in each 
cluster of the motion flow model are shown in Figure~\ref{fig:grasploc}.
Interestingly, the robot successfully reached and grasped the object 
in the regions, where the interacting demonstrations have not covered 
(see the green cylinders).  
However, if the location of the object is too far from the collected
demonstrations, it fails (see the red cylinder).

\begin{figure*}[t!]
	\centering
	\subfigure[]{\includegraphics[width=1.3\columnwidth]
		{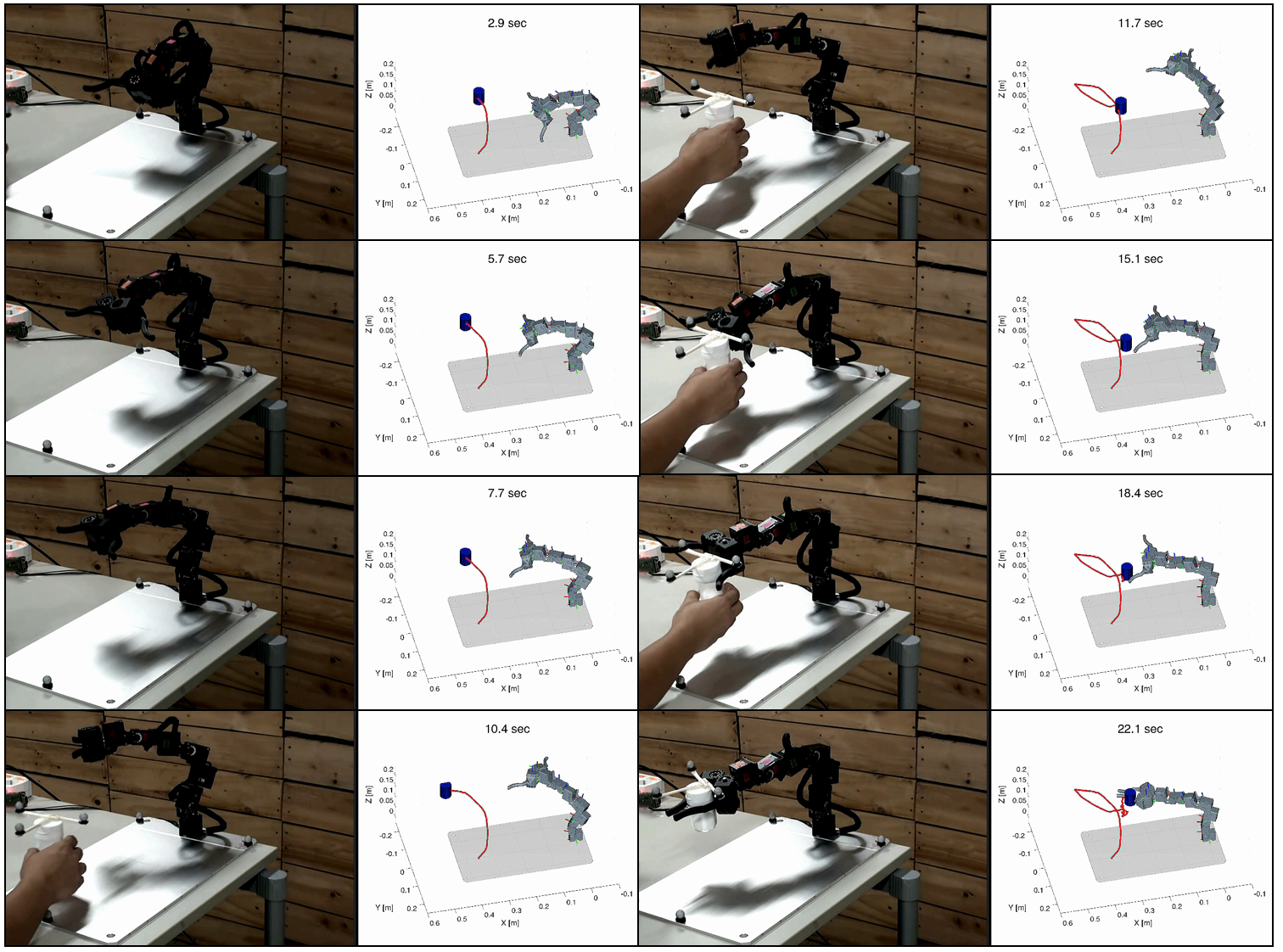}
	\label{fig:intention_a}}
	\hspace{2em}%
	\subfigure[]{\includegraphics[width=0.645\columnwidth]
		{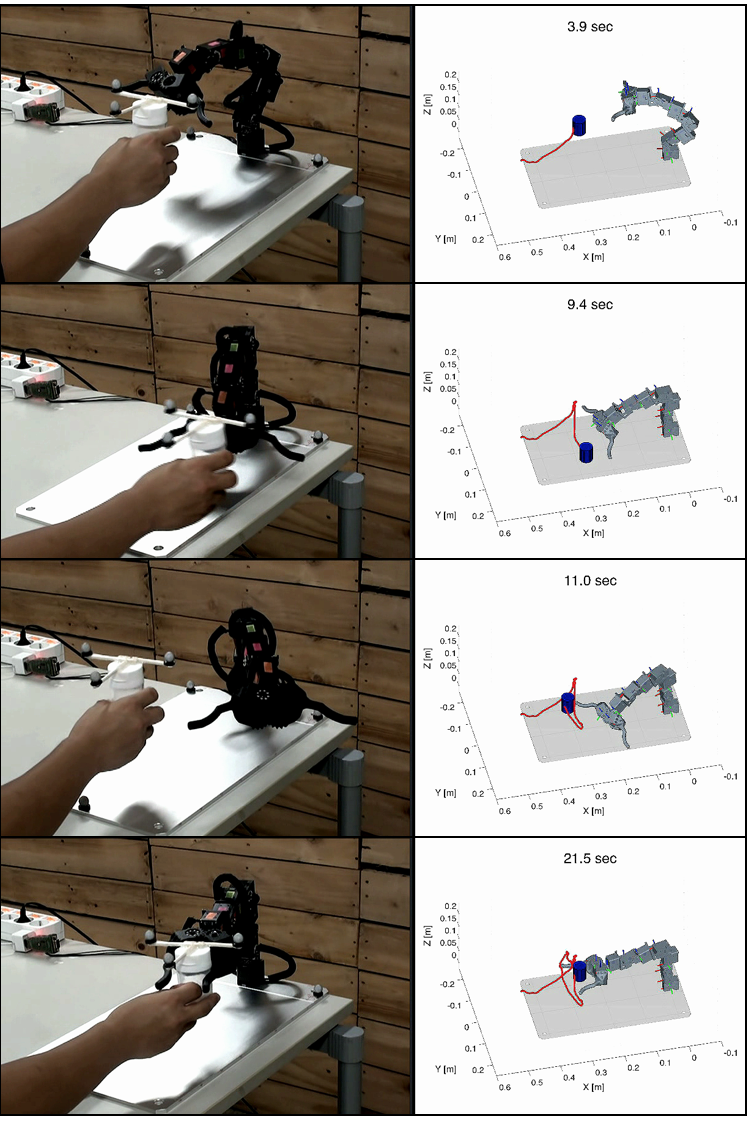}%
	\label{fig:intention_b}}
	\caption{
		Snapshots of experiments using Manipulator-X. 
		(a) The intention is changed from 
		left-swipe to center-hand-over. 
		(b) The intention is changed from 
		center-hand-over to right-hand-over and back to center-hand-over. 
	}
	\label{fig:intentionchange}
\end{figure*}

The objective of the second set of experiments
is to see how the robot reacts to varying intentions of a human. 
As the trajectory of Manipulator-X is re-planned 
every two seconds, it should be able to correctly modify the trajectory
based on recognized motions. 
Figure \ref{fig:intentionchange} shows two cases: 
(1) the human co-worker changes his intention from left-swipe to
center-hand-over; and 
(2) the intention is changed from center-hand-over to right-hand-over
and back to center-hand-over.  
It shows that the proposed IAAL algorithm can successfully recognize
the change in the user's motions and respond accordingly.

\section{Conclusion}

In this paper, a motion flow model that describes
motion trajectories of a human is proposed
for human robot cooperation tasks. 
The main contribution of this paper is the nonparametric 
motion flow similarity measure based on both spatial and temporal 
similarities using a Gaussian process. 
The time-alignment of trajectories is not required for the proposed
similarity measure and it plays an important role in early recognition
of human motions. 
We presented a human robot cooperation algorithm 
based on the proposed motion flow model
and compared it with a mixture of interaction primitives algorithm.
The proposed algorithm has shown a superior performance 
with respect to the prediction error when partial trajectories of a
human coworker are given. 

\bibliographystyle{IEEEtran}
{ \small
\bibliography{bib_iros,HAndyParkBib2016,relatedwork,references}
}

\end{document}